\documentclass[12pt]{article}

\usepackage{graphicx}
\usepackage{amsmath}
\usepackage{amsthm}
\usepackage{amssymb}
\usepackage{enumerate}

\usepackage{natbib}

   \newtheorem{theorem}{Theorem}[section]

\usepackage{bm}
\usepackage{amsmath}
\usepackage{url}
\usepackage{amssymb}
\usepackage{graphicx}
\newcommand{\be}{\begin{equation}}
\newcommand{\ee}{\end{equation}}
\newcommand{\bea}{\begin{eqnarray}}
\newcommand{\eea}{\end{eqnarray}}

\newcommand{\la}{\lambda}

\def\<{\langle} 				
\def\>{\rangle} 				

\newcommand{\beq}{\begin{equation}}
\newcommand{\eeq}{\end{equation}}
\newcommand{\bbar}{\begin{eqnarray}}
\newcommand{\eear}{\end{eqnarray}}

\begin{document}

\begin{center}
{\Large \bf  When learners surpass their models: mathematical modeling of learning from an inconsistent source}
\end{center}

\begin{center}
{\large \bf Yelena Mandelshtam and  Natalia L. Komarova
}
\end{center}

\noindent Department of Mathematics, University of California Irvine, Irvine CA 92697



\paragraph{Abstract.} We present a new algorithm to model and investigate the learning
         process of a learner mastering a set of grammatical rules
         from an inconsistent source. The compelling interest of human
         language acquisition is that the learning succeeds in
         virtually every case, despite the fact that the input data
         are formally inadequate to explain the success of
         learning. Our model explains how a learner can successfully
         learn from or even surpass its imperfect source without
         possessing any additional biases or constraints about the
         types of patterns that exist in the language. We use the data
         collected by \cite{singleton2004learners} on the performance of a 7-year boy Simon, who
         mastered the American Sign Language (ASL) by learning it from
         his parents, both of whom were imperfect speakers of ASL. We
         show that the algorithm possesses a frequency-boosting
         property, whereby the frequency of the most common form of
         the source is increased by the learner. We also explain
         several key features of Simon's ASL.

\bigskip 


\section{Introduction}

The ability of children to ``improve'' the language of their parents
has been widely documented. One famous example comes from the studies
of the origins and development of the Nicaraguan Sign Language
\citep{senghas1995development, senghas1997argument,
  senghas2001children}. This language was spontaneously developed by
deaf children in a number of schools in Western Nicaragua in the 1970s
and 1980s. Further, the phenomenon of creolization of pidgin languages
has been studied \citep{andersen1983pidginization,
  thomason1991language, sebba1997contact}. In the course of one
generation, a language is created from a multilingually-derived, limited communication
system, where different people only have rudimentary knowledge of each
other's native languages. 

It is argued that humans have an ability to improve on the (possibly inconsistent) linguistic input that they receive. The phenomenon of language regularization has been studied in the context of historical linguistics \citep{kroch1989reflexes, kroch1997verb, pearl2007input}. Language over-regularization in children has also received much attention \citep{marcus1992overregularization, marcus1995children, marchman1997overregularization, plunkett1999connectionist}.

A unique example of a language regularization by a single child is
reported in \cite{singleton2004learners}. In this paper, the authors
analyzed the language of a deaf boy (whom they named Simon). The unique
situation of Simon was that his parents were his only sources of
American Sign Language (ASL). Because Simon's parents were not native
speakers of ASL (they both learned it after the age of 15), the
language that they spoke to Simon had many inconsistencies. However,
in a study of Simon, his parents, and other native deaf children,
Simon greatly outperformed both of his parents, although he was still
somewhat behind the native speakers in certain aspects.

The phenomenon of language regularization showcased by Simon was
studied in depth by Elissa Newport and her colleagues, in a number of
elegant experiments performed with adult and children learners. In
\cite{kam2009getting} and \cite{hudson2005regularizing}, artificial
miniature language acquisition from an inconsistent source was
studied.  The participants learn the language by listening to
sentences of the language, which are presented in an inconsistent
fashion (allowing for a probabilistic usage of several forms). The
structure and complexity of the probabilistic input varies from
experiment to experiment. The goal is to assess what kinds of input
are most consistent with the tendency of adults and children to
regularize. The authors also evaluate the differences in the learning
patterns between adults and children. It was shown that
children achieved higher degrees of regularization than adults, and
that the degree of regularization varied in a predictable fashion
depending on the structure of the source.

Newport and colleagues used the terms ``frequency boosting'' and
``frequency matching'' to describe the amount of regularization
exhibited by learners.  Let us suppose that the ``teacher'' (or the
source of the linguistic input) is inconsistent, such that it
probabilistically uses several forms of a certain rule. Frequency
boosting is the ability of a language learner to increase the
frequency of usage of a particular form compared to the
source. Frequency matching happens when the learner reproduces the
same frequency of usage as the source.

Mathematical modeling of frequency boosting has been performed by
\cite{reali2009evolution}. By using the iterated learning model (see
\cite{kirby1999function, kirby2001spontaneous,
  brighton2002compositional, smith2003iterated, kirby2004ug}) with
rational Bayesian agents \citep{griffiths2007language}, it was shown
that language regularization can be achieved in the course of several
generations of learners. This is an important mechanism that has also
been documented by \cite{smith2010eliminating}, which studied the
gradual, cumulative population-level processes giving rise to language
regularity.

In this paper we provide a mathematical framework that allows us to
model regularization achieved within a single generation. We create a
model (of the reinforcement-learner type) that allows us to study
Simon's learning behavior, as well as other situations of language
regularization in adults and children. 

This rest of the paper is organized as follows. In Section
\ref{sect:lear} we motivate the discussion of learning algorithms, and
introduce a particular algorithm of the reinforcement type. Section
\ref{sect:prop} studies properties of this algorithm analytically and
numerically. In particular, its frequency boosting property is
demonstrated, and the speed of convergence investigated. Section
\ref{sect:sim} uses the algorithm to discuss the performance of Simon
in \cite{singleton2004learners}. Section \ref{sect:disc} contains
discussion and conclusions.

\section{The learning alorithm}
\label{sect:lear}

\subsection{Learning algorithms, a description of the concept}
   In creating a mathematical model, it is assumed that the learner has
to master a number of different grammatical rules. It can be further
assumed that they are independent of each other, that is, that the
learning of one rule does not depend on the state of knowledge of the
rest of the rules (this is a simplification). Because of the latter
assumption, different rules will be considered separately, which
simplifies the picture. 

This study will concentrate on the process of learning of one rule. In
general, each rule could have multiple variants (forms). The input in
this context is a number of applications of the rules. Of course,
total linguistic input will contain sentences which do not contain
applications of the given rule. Such sentences do not contain any
information regarding the rule, and thus should be ignored. Thus the
relevant part of the input can be presented as a string of numbers,
each number representing a given form being used. So, the $i$th number
written in the string of numbers represents the $i$th form that the
learner was exposed to. For example, if a given rule has two forms, a
particular linguistic input may be 1212222212112, that is, the first
time the learner was exposed to form 1, then the second time to form
2, etc. The task of the learner is to evaluate the input and create an
output, which reflects the input in some way.

There may be different algorithms which lead to different results of
learning, even if the same input is used. An example is the following
algorithm (Algorithm A1): “Consider the first application of the rule,
and ignore all the rest of the information”. If this algorithm is
implemented, the learner will always learn the form of the rule that
(s)he first encounters. This is not a realistic algorithm, but an
example of a valid, well defined learning procedure.

Another example is Algorithm A2: “Use all the input received. Count
how many times forms 1 and 2 were used. Then use forms 1 and 2
randomly, with the same proportion.”

Algorithms A1 and A2 defined above are in some sense opposites of each
other. Algorithm A1 leads to perfect consistency of learning, that is,
the learner will always use the same rule, regardless of the degree of
inconsistency of the source. Algorithm A2 will retain the degree of
inconsistency of the source. Also, the two algorithms have very
different computational requirements for the learner: Algorithm 1 only
needs to retain the information from the first application of the
rule, while Algorithm 2 needs to remember and analyze a whole string
of input. 

Algorithm A1 has a serious flaw. Consider the following input:
“2111111111111”, that is, the first sentence contains form 2, and the
rest will contain form 1. Algorithm A1 will result in form 2, whereas
the source is almost perfectly consistent (except for one application)
in using form 1. Since one of the goals is to learn a language
somewhat close to that of the source, Algorithm A1 does not do a good
job. It learns (consistently) a wrong rule. It can also be seen in the
following way: Algorithm A1 is not robust in terms of slight errors of
the source. Even if the source is completely consistent, the algorithm
must allow for occasional errors (or imperfections in
communication). The first application of the rule in “2111111111111”
could be just the result of bad communication (noise). Algorithm A1
has no way to buffer that. Algorithm A2 does not have this problem,
because for a sufficiently long string of input, errors will make a
negligible effect on the result.

Neither of the algorithms exhibit a frequency boosting property
consistently. That is, Algorithm A1 ``boosts'' the frequency of the
form that happened to be used first by the source (which may or may
not correspond to the more frequent form). Algorithm 2 has a frequency
matching property by construction.

In oder to describe the frequency boosting/regularization behavior observed in human learners, it is desirable to design a learning algorithm which meets the following intuitive
requirements:
\begin{enumerate}
\item Learn the rule which is close to that of the source. If the source is consistent, then it should be the same rule with a high probability.
\item Improve the consistency of the source. 
 \item Be robust with respect to errors of communication/noise. 
 \item Be computationally inexpensive.
\end{enumerate}
In the next section we introduce an algorithm with these properties. 

\subsection{Formulation of the algorithm}
\label{sect:alg}

The key idea of the algorithm presented in this work is that the
states that are being modified by the probabilistic input are
themselves probabilities. 

Suppose a rule only has two forms, form $1$ and form $2$. At each
instance of time, the learner is characterized by two numbers, $X_1 \ge 
0$ and $X_2 \ge  0$, each corresponding to probability of the usage of the corresponding
form of the rule (we further impose $X_1+X_2>0$). In particular, the probability of the learner
to use form $1$ is given by $X_1/(X_1 + X_2)$, and the probability of the
learner to use form $2$ is given by $X_2/(X_1 + X_2)$. One can see that if $X_1
= 0$, then form $2$ is always used (the learner's language is consistent
with respect to the rule in question). Similarly, with $X_2 = 0$, only
form $1$ is used. 

Each time an instance of the rule application is received, the learner
updates the values $X_1$ and $X_2$. There are many ways in which this
can be done, see reinforcement-learning models
\citep{sutton1998reinforcement, norman}. Reinforcement
models have played an important role in modeling many aspects of
cognitive and neurological processes, see
e.g. \cite{maia2009reinforcement, lee2012neural}. This work presents a novel algorithm that satisfies regulations outlined above. Let us set $X_1 +
X_2 = L$, where $L$ is a given positive integer. If form $1$ was used by the source, then the learner updates in the following way: if $X_1 < L$ then
$$X_1 \to X_1 +1,	X_2 \to X_2 -1,$$
otherwise, no change $(X_1 \to X_1, X_2 \to X_2)$. If form $2$ was used, then the learner
updates as follows: if $X_2 < L$ then 
$$X_1 \to X_1 -1,	X_2 \to X_2 +1,$$
otherwise no change. 

This algorithm can be easily generalized to several forms of the same
rule. Suppose that there are in total $M$ different ways in which the
rule can be used (forms $1,...,M$). Then, at each instance of time,
the learner is characterized by a vector of $M$ nonnegative numbers,
${\bf X} = [X_1,...,X_M]$, such that $\sum_{i=1}^M {X_i}=L$; we say
that the vector ${\bf X}/L$ belongs to an $M$-dimensional
simplex. After each exposure to the rule usage, the values $X_i$ are
updated. If form $j$ is used by the source, the learner applies the
following update rule:
\begin{align}\text{if}\,\,i &\not= j, X_i \to X_i-\delta_i^-, \delta_i^- =
\begin{cases}
\frac{s}{M-1},\,\mbox{if } X_i > \frac{s}{M-1}\\ X_i, \,\mbox{otherwise}
\end{cases}\notag\\
\text{if} \,\,  i&=j, X_j \to X_j + \delta^+, \delta^+=
\displaystyle\sum\limits_{i\not=j}\delta_i^-.\end{align}
Here the parameter $0< s < L$ defines the increment of a learning
update. Clearly, multiplying $s$ and $L$ by the same number does not
change the algorithm, so it will be assumed that $s=1$.

\section{Properties of the learning algorithm}
\label{sect:prop}

An important application in our context is to consider a
teacher-learner pair. In general, the teacher does not necessarily
have to be one person, it could be a number of people. The only
requirement is that the statistics of the source do not change in
time. To be more precise, suppose that the source is characterized by
nonnegative numbers $\nu_1, ... , \nu_M$, with $\sum_{i=1}^M\nu_i=1$. These
are interpreted as probabilities of the source to emit each of the $M$
forms of the rule. Thus, the source can be thought of as being just
one agent (whose values of $X_i$ are ``frozen,'' that is, not
updated), or it could be a collection of such non-updating agents.
The situation with the teacher-learner pair can be used to model the
language acquisition of a child from one or two parents and other
adults whose language is frozen. In particular, it can be applied to
the situation of Simon who learned only from his two parents. It is
also directly applicable to the experiments in which adults or children
are learning from an artificial/inconsistent source, such as in papers
\cite{kam2009getting, hudson2005regularizing}.

\subsection{Analytical Results}

Let us consider the case $M=2$, or two alternative forms of the
rule. We denote by $\nu$ the probability of the source to use form 1;
form 2 is used with probability $1-\nu$.  The reinforcement learner
algorithm described above can be studied as a Markov chain with states
$0, 1/L, 2/L...1$, where the constant $L$ appears in the description
of the algorithm, Section \ref{sect:alg}. The states are
probabilities, that is, if the Markov Chain is in state $i/L$, the
probability that the learner will use form 1 is $i/L$. This
chain can be modeled by the distribution vector $[X_0, X_1, ...,
  X_L]$, at any point in time, where $X_i$ represents the probability
that the chain is in state $i/L$. The distribution vector is updated
from step to step by multiplying by the transition matrix, $A$. In
this case, the transition matrix will only have elements $\nu$ - the
probability that the source uses form 1, meaning that the
chain will move to the next state, $1-\nu = \mu$ - the probability that
the source uses form 2, meaning that the chain will move
to the previous state, and $0,$ since for all other cases, it is
impossible to move directly from one state to another one. We have
$$
A=
\begin{bmatrix}
\mu & \nu & 0 & \cdots & \cdots & 0\\
\mu & 0 & \nu & 0 & \cdots& 0\\
0 & \mu & 0 & \nu & \cdots & 0\\
\vdots & 0 & \ddots & 0 & \ddots & \vdots\\
\vdots &    &   & \ddots& 0 & \nu\\
0 & \cdots & 0 & 0 & \mu & \nu
\end{bmatrix}
$$
Let $\Pi$ be the stationary distribution vector of the matrix $A.$ No
matter what the original distribution was, as the number of steps goes
to infinity, the vector will become infinitely close to the stationary
distribution.

Call $P$ the weighted average of the states, or the probability that a
learner will use the correct form  after a very large number of sentences
given by the source. $P$ can be found by the formula 
\begin{equation}\label{average}P(L, \nu) = \displaystyle\sum_{i=0}^L
{\frac{i}{L}\Pi_i}.
\end{equation}

We have the following result. 
\begin{theorem} For all $L$, the expected frequency of the learner in the quasi-steady state is given by
\beq
\label{P}
P(L,\nu)=1+\frac{1}{L}\left(\frac{L+1}{\lambda^{L+1}-1}-\frac{1}{\lambda-1}\right),
\eeq
where $\lambda=\nu/(1-\nu)$.
\end{theorem}

{\bf Proof}: In order to find $P$, it is necessary to find $\Pi$
first. The stationary distribution $\Pi$ satisfies
$$ 
[\Pi_0, \Pi_1, \Pi_2....,\Pi_L]
\begin{bmatrix}
\mu & \nu & 0 & \cdots & \cdots & 0\\
\mu & 0 & \nu & 0 & \cdots& 0\\
0 & \mu & 0 & \nu & \cdots & 0\\
\vdots & 0 & \ddots & 0 & \ddots & \vdots\\
\vdots &    &   & \ddots& 0 & \nu\\
0 & \cdots & 0 & 0 & \mu & \nu
\end{bmatrix}   = [\Pi_0, \Pi_1, \Pi_2....,\Pi_L]
$$
which is equivalent to a system of equations:
$$
\begin{cases}
\mu\Pi_0+\mu\Pi_1=\Pi_0\\
\nu\Pi_0+\mu\Pi_2=\Pi_1\\
\vdots\\
\nu\Pi_{L-2}+\mu\Pi_L=\Pi_{L-1}\\
\nu\Pi_{L-1}+\nu\Pi_L=\Pi_L
\end{cases}
$$
The first equation is solved to get 
$\Pi_1=\frac{1-\mu}{\mu}\Pi_0$.
Because $1-\mu=\nu,$ this can be re-written as: $\Pi_1=\frac{\nu}{\mu}\Pi_0$.
Using the new value for $\Pi_1$, the next equation is solved to
obtain $\Pi_2=\frac{\nu-\nu\mu}{\mu^2}\Pi_0$, which can also be written as
$\frac{\nu^2}{\mu^2}$.\\
To solve for the general term, induction is used with
base case $\Pi_1=\frac{\nu}{\mu}\Pi_0.$ 
Assume $\Pi_i=\frac{\nu^i}{\mu^i}\Pi_0$ and
$\Pi_{i+1}=\frac{\nu^{i+1}}{\mu^{i+1}}\Pi_0,$
then 
$$\nu(\frac{\nu^i}{\mu^i})\Pi_0+\mu(\Pi_{i+2})=\frac{\nu^{i+1}}{\mu^{i+1}}\Pi_0.$$
This is solved to get: 
$$\Pi_{i+2}=\frac{\nu^{i+2}}{\mu^{i+2}}\Pi_0.$$
So, by induction, each term of the stationary distribution has the form,
$$\Pi_m=\frac{\nu^m}{\mu^m}\Pi_0,\quad  m= 0, ... , L.$$ 
The last equation in the system is used to confirm this.

Therefore, the stationary state of the matrix is:
\begin{equation}
\label{stat}
[\Pi_0,
\frac{\nu}{\mu}\Pi_0, \frac{\nu^2}{\mu^2}\Pi_0, ... , \frac{\nu^L}{\mu^L}\Pi_0],
\end{equation}
which is also the distribution of probabilities for the learner to be in state 
values $0/L,1/L ... , L/L$  as
$n\mbox{(the number of steps)}$ goes to $\infty.$

Because the vector is a distribution of probabilities, its components add up
to one. Thus, it is possible to solve for $\Pi_0$:
$$\Pi_0 + \frac{\nu}{\mu}\Pi_0+ ... + \frac{\nu^L}{\mu^L}\Pi_0=1$$
or
$$\Pi_0(\frac{1-\frac{\nu}{\mu}^{L+1}}{1-\frac{\nu}{\mu}})=1.$$
It is convenient to introduce a new parameter, $\la=\frac{\nu}{\mu}$, obtaining
\begin{equation}
\label{pii}
\Pi_0=\frac{1-\la}{1-\la^{L+1}}
\end{equation}
Now, to find the weighted average, $P(L,\nu)$, it is necessary to use
formula (\ref{average}).
By (\ref{stat}), (\ref{pii}), it is equal to:
$$P(L,\nu)=\frac{1-\la}{L(1-\la^{L+1})}\displaystyle\sum_{i=0}^L{i\la^i}.$$
Solve and obtain equation (\ref{P}). 
\qed
\bigskip 

The following theorem rigorously establishes the boosting effect in the case with
two forms.
\begin{theorem} For all $L$ and $\nu>1/2$, $P(L, \nu)>\nu.$
\end{theorem}

{\bf Proof}: We need to prove that the expression in (\ref{P}) is greater than $\nu$, that is, that $P(L,\nu)/\nu>1$. 

Note that $\nu$ can be expressed through $\la$ as
$\nu=\frac{\la}{1+\la}.$ Therefore, it is sufficient to prove the inequality:
$$1+\frac{1}{L}\left(\frac{L+1}{\lambda^{L+1}-1}-\frac{1}{\lambda-1}\right)>\frac{\la}{1+\la},$$
which, with some algebraic manipulation, taking into consideration the
fact that $\la>1$, can be transformed into 
$$L\la^{L+1}+\la+L\la+1 > L+L\la^{L}+\la^{L}+\la^{L+1}. $$
Now, introduce functions
$$f(\la)=L\la^{L+1}+\la+L\la+1 \mbox{ and } g(\la)=L+L\la^{L}+\la^{L}+\la^{L+1}.$$
It follows that $f(1)=2L+2=g(1)$, and also
$f'(\la)=L(L+1)\la^{L}+1+L$ and
$g'(\la)=L^{2}\la^{L-1}+L\la^{L-1}+(L+1)\la^L.$
When $\la=1$, we have $f'(\la)=(L+1)^2=g'(\la).$
The second derivatives of $f$ and $g$ are:
$$f''(\la)=L^2(L+1)\la^{L-1} \mbox{ and }
g''(\la)=L^2(L-1)\la^{L-2}+L(L-1)\la^{L-2}+L(L+1)\la^{L-1}.$$
Since by definition the state space ($L$) is greater than one, and
$\lambda >1$,  it follows that $(L-1)\la >L-1$.  Thus
$L^2(L+1)\la^{L-1}>L(L+1)(L+\la-1)\la^{L-2}$, which implies $f''(\la)>g''(\la)$ for $\la>1$.

Therefore, because $f''(\la)>g''(\la)$ for $\la>1$, and $f'(1)=g'(1)$ and
$f(1)=g(1)$ we have $f(\la)>g(\la)$
for all $\la>1$, by the fundamental theorem of calculus. This means
that 
$$P(L,\nu)>\nu,\quad \nu>1/2,$$
that is, the algorithm has the boosting property.  
\qed

\bigskip 

The following theorem shows that as the state space gets sufficiently
large, the boosting effect is maximized, resulting in a perfectly
consistent output.
\begin{theorem} For all $\nu>1/2$, $\displaystyle\lim_{L\to\infty}P(L,\nu)
  =1.$
\end{theorem}

{\bf Proof:} By evaluating the limit as $L\to \infty$ in equation (\ref{P}) by standard methods under the assumption that $\lambda>1$, we obtain the desired result. 
\qed

\subsection{Numerical Simulations}

The reinforcement learner algorithm contains only one parameter, $L$. It is more intuitive to talk about the quantity
$$s=\frac{1}{L},$$
the increment of learning. This is the amount by which the
probabilities change at each step, following the source's input. The
analytical results reported in the previous section can be summarized
as follows. In the case where $M=2$ (two forms of the rule), as the
number of steps increases, the learner converges to a quasi-stationary
state, where it is characterized by the frequency of form 1 given by
equation (\ref{P}). This frequency is higher than the frequency of the
source, $\nu$ (the frequency boosting effect). The frequency boosting
effect becomes more pronounced as $s\to 0$, that is, for small values
of the increment.

  \begin{figure}
  \centering 
  \includegraphics[scale=0.4]{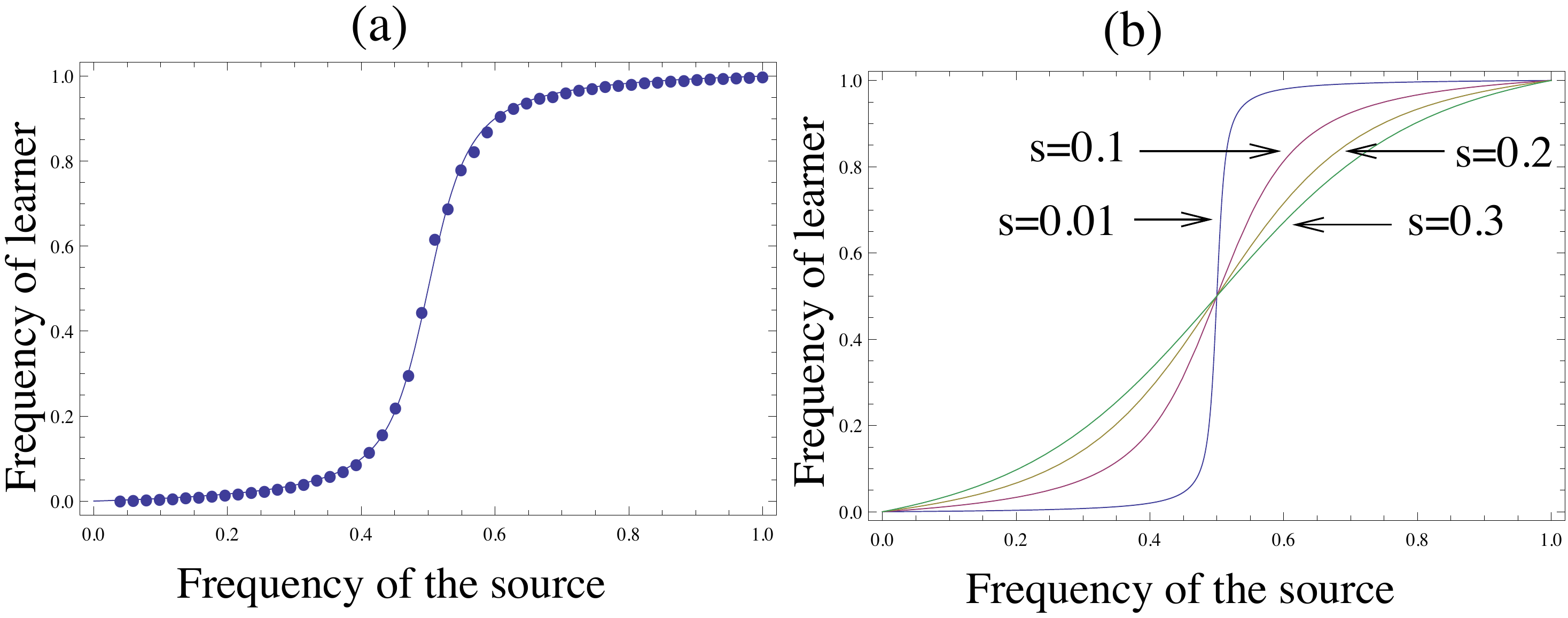}
   \vspace{\baselineskip}    
 \caption{\footnotesize The boosting property of the algorithm for $M=2$ forms. (a) The agreement of formula (\ref{P}) (the solid line) with numerical simulation of the expected output frequency of the learner after $30,000$ iterations (the dots). The expected frequency of the learner is plotted against the frequency of the source, $\nu$, for $s=0.05$. (b) The dependence of the boosting propertty on the update parameter, $s$. The expected frequency of the learner is plotted as a function of the frequency of the source for four different values of $s$.  }
\label{fig:boo}
\end{figure}

\begin{figure}
 \centering 
\includegraphics[scale=0.25]{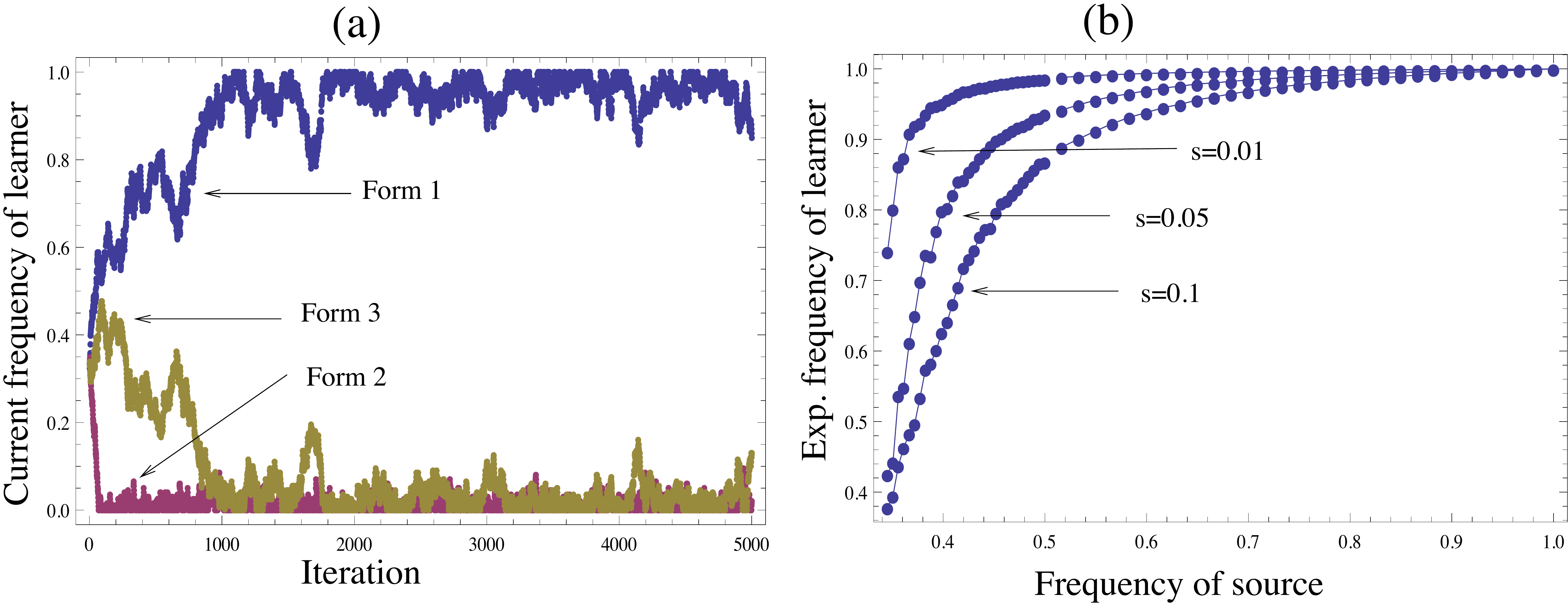}
   \vspace{\baselineskip}    
   \caption{\footnotesize The boosting proerty of the algorithm for $M=3$ forms. (a)  A typical numerical run is shown where the current frequencies of the learner for forms 1, 2, and 3 are plotted against the iteration number. The frequencies of the source are $(\nu_1,\nu_2,\nu_3)=(0.4,0.25,0.35)$, and $s=0.01$. (b) The dependence of the boosting propertty on the update parameter, $s$.  The expected frequency of the learner for form 1 is plotted against the frequency of the source, $\nu_1$, for three different values of $s$. The frequency of the source is taken to be $(\nu_1,\nu_2,\nu_3)=(\nu_1,(1-\nu_1)/2,(1-\nu_1)/2)$.}
\label{fig:mor}
\end{figure}

To find out numerically whether the reinforcement learner algorithm
possesses a source boosting property for $M>2$, and also to
investigate other properties of the algorithm, probability vector
components of the learner were computed numerically as a function of various
parameters. Repeated teacher-learner interactions were simulated, and
after each interaction, the current probability vector of the learner
was recorded. A Fortran computer code written computed the {\it output
  frequency}, i.e. the probability that the learner would use the
correct form. The output frequency was computed as a function of time
(the number of ``sentences'' learned from the source), the size of the
increment, and input frequency.

Figure \ref{fig:boo}(a)  shows the agreement of the analytical prediction  (\ref{P}) with the numerical simulations, in the case of two forms, $M=2$. We can see from figure \ref{fig:boo}(b) that as $s$ increases, the boosting property becomes weaker. Figure \ref{fig:mor} numerically demonstrates the existence of the boosting property for higher numbers of $M$.  As in the $M=2$ case, the frequency boosting effect becomes more pronounced as $s\to 0$.

\begin{figure}
   \centering 
   \includegraphics[scale=0.25]{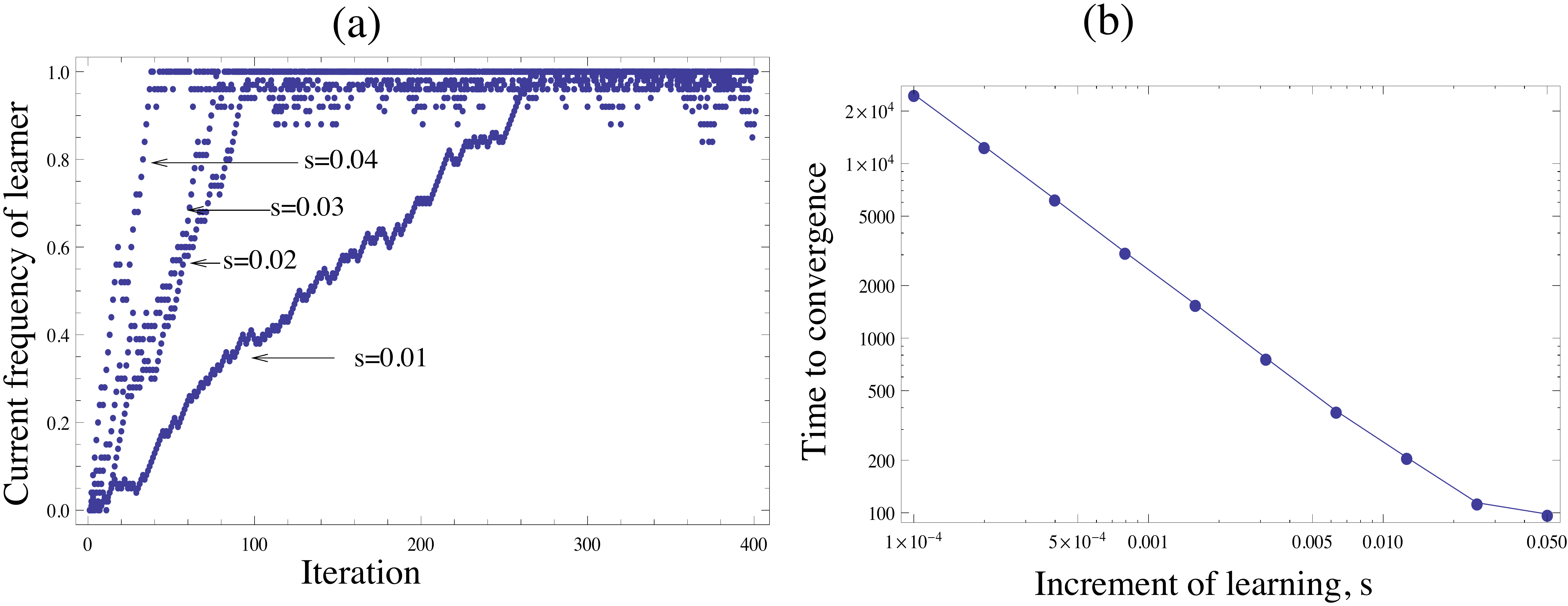}
    \vspace{\baselineskip}    
    \caption{\footnotesize Convergence time of the algorithm for $M=2$ forms. (a) Typical numerical runs are shown where the current frequency of the learner is plotted against the iteration number, for four different values of $s$. (b) The time to convergence is plotted as a function of the quantity $s$. This was calculated as the number of iterations until the moving average frequency of the learner reached the predicted mean frequency (formula (\ref{P})) within the margin of $0.1\%$. The moving average was taken over $200$ iterations, and the resulting convergence times averaged over 200 runs.  The frequency of the source for (a) and (b) is $\nu=0.7$.}
 \label{fig:tim}
 \end{figure}

\begin{figure}
   \centering 
   \includegraphics[scale=0.22]{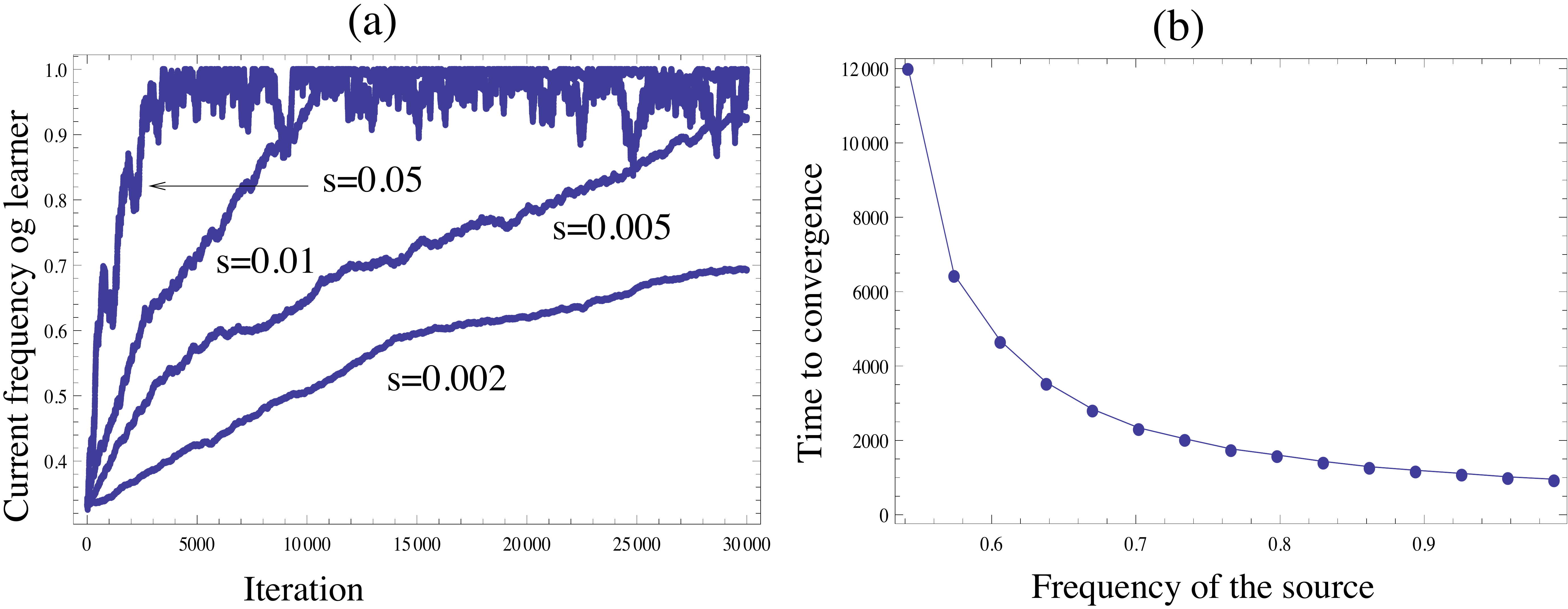}
    \vspace{\baselineskip}    
    \caption{\footnotesize Further properties of the learning algorithm. (a) For a multi-form source with $M=3$, the typical numerical runs are shown where the current frequency of the learner for form 1 is plotted against the iteration number, for four different values of $s$. The frequencies of the source are $(\nu_1,\nu_2,\nu_3)=(0.4,0.25,0.35)$. (b) Convergence time of the algorithm for $M=2$ forms as a function of the frequency of the source. The increment was taken $s=0.001$. This plot was produced by taking the miving average, and then averaging over 200 runs, as in figure \ref{fig:tim}(b).}
 \label{fig:con}
 \end{figure}

Next, the dependence of the convergence time on the input frequency
was studied. Figure \ref{fig:tim}(a) shows some typical runs for
several values of the learning increment, $s$. It is clear thas as $s$
increases, the algorithm converges faster. This was investigated
formally in figure \ref{fig:tim}(b). The convergence time was
calculated by computing output probabilities taking a running average
over $200$ steps to smooth out the curve, and then by defining the
time to converge as the first step, $n$, such that the output
frequency at $n$ steps is within $.001$ of the analytically calculated
expected output frequency. All of the data was averaged over 200
trials. To investigate how the convergence time depends on the
increment size, figure \ref{fig:tim}(b) plots the convergence time as
a function of $s$. We can see that convergence time decreases as $s$
increases. The same trend holds for multi-form inputs, see figure
\ref{fig:con}(a) where we plot the current frequency of learners for
$M=3$, for four typical runes with different increments $s$. Figure \ref{fig:con}(b) plots the convergence time as a function of the input
frequency for $M=2$. It decreases with increasing input frequency.

To summarize,  as the increment $s$
decreases, the boosting effect increases, but the speed of convergence
decreases. Further, the speed of convergence increases with the input frequency.

\section{Applications: Simon}
\label{sect:sim}

The numerical and the analytical studies both showed that the
reinforcement learner algorithm possesses a source boosting
property. In particular, when there are two forms, the learner boosts
the one that is used more than $1/2$ of the time, and more generally,
if there are $M$ forms, the learner boosts the dominant form, even if
it is used with a frequency barely greater than $1/M$. This result
explains that it is possible for a learner to surpass its source
without possessing any innate sense of grammar. The study of the
dependence of the convergence time on the input frequency or the state
space allows to determine how many input sentences are required to
achieve the boosting under various learning scenarios. These
observations can help one to explain certain learning phenomena, such
as the reason why those learning from an inconsistent input learn
more slowly than those learning from a consistent input, and by how much
the speed is slowed down.

\begin{figure}
  \centering 
   \includegraphics[scale=0.25]{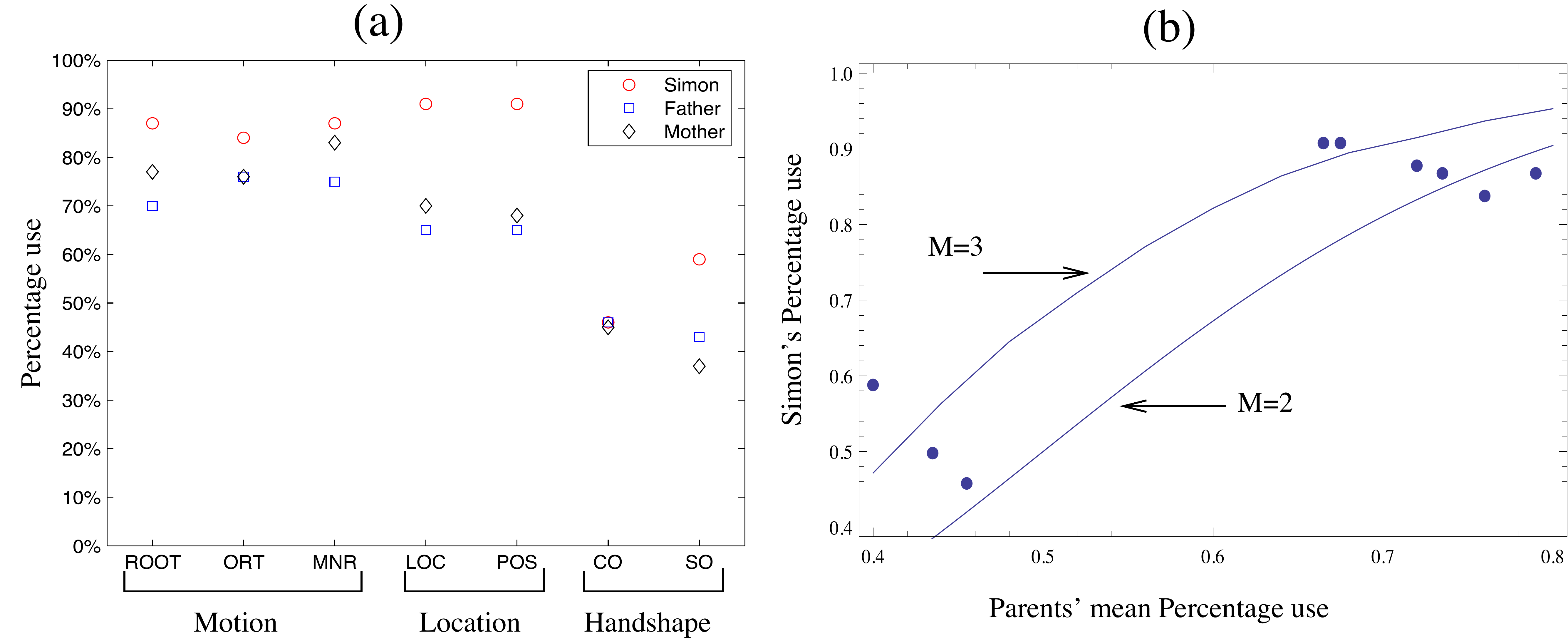}
    \vspace{\baselineskip}    
    \caption{\footnotesize (a) The performance of Simon and his parents
      (percent correct score) for seven morpheme categories: Root,
      Orientation, Manner, Location, Position, Central Object and     Secondary Object), data from \cite{singleton2004learners}. The
      morphemes belong to three groups, Motion, Location, and
      Handshape, as indicated. (b) The same data are redrawn to show the boosting effect (the dots). The two solid lines represent the best fit for the model with $M=2$ and with $M=3$.}
 \label{fig:sim}
 \end{figure}

In particular, these results are in agreement with the observational
studies of \citep{singleton2004learners}. Simon was tested on $7$
movements, which were split into 3 categories: handshape, motion, and
location. Figure \ref{fig:sim}(a) presents some results from Singleton
and Newport's study on Simon's performance compared to that of his
parents, categorized by different morphemes (different rules). Simon's
results surpassed those of his parents in both motion and location
(which were both in the 65-75\% range) by about 20\%, boosting the
score to the score of the native learners. Thus, Simon greatly
outpreformed his parents, even though they were his only sources. This
can be explained by the boosting effect described here. Figure
\ref{fig:sim}(b) presents the same data as in part (a) of the figure,
plotted in a different way. Percentage use for the parents (the
horizontal coordinates of the dots) were computed by taking the mean
of the mother's and the father's frequency of use (that is, we asumed
that Simon was equally exposed to both parents' input). The vertuical
coordinate is Simon's percentage of use. The two solid lines represent
the fits from our model based on $M=2$ and $M=3$ forms. Although these
fits show that the theory can give results roughly in the ball park of
the observations, we must note that the data shown in figure
\ref{fig:sim} are averages of performance over several morphemes in
each class, and further data need to be used to inform more precise
model parameterization. It is more instructive to look at individual
examples of Simon's performance, which is done next.

An interesting question that was addressed in the paper was why Simon's
performance was characterized by a high degree of frequency boosting
for motion and location morphemes, but his frequency boosting was much
more modest and even nonexistent for handshape morphemes. Several
hypotheses were suggested. Here we provide some mathematical
foundation for those ideas.

One hypothesis was that ``Simon's parents do not always use the target
ASL form as their most frequent response''. It is clear from figure
\ref{fig:sim}(a) that for Motion/Location morphemes, the parents' most
frequent rule variant was indeed the correct one, consistent with the
standard ASL. A frequency boosting as described in the previous
sections would increase the frequency of that form, leading to a
larger percentage of correct usage for Simon. This is consistent with
the observations.

On the other hand, for handshape morphemes, the situation appears to
be more complex. From figure \ref{fig:sim}(a) we can see that with
both types of handshape morphemes, the parents' performance was
particularly poor in terms of the frequency of usage of the correct
ASL morphemes. But Simon's performance is very different for Central
Object (CO) and Secondary Object (SO) morphemes. In the case of CO
morphemes, Simon does not appear to boost frequency at all, while for SO we can see a considerable degree of frequency boosting. 

To explain this, we will first take a closer look at individual CO morphemes. It turns out that because both parents use the correct form less than $59\%$ of the time, Simon's 
frequency boosing in this case  resulted in a decrease in the
frequency of the correct variant, and instead in the boosting of an alternative,
incorrect form. This made Simon's usage more consistent than his
parents', but not more correct in the sense of being closer to the
standard ASL. The parents's most consistent sign for ``vehicle'' was the
so-called ``B-edge' which was an incorrect form for ASL, see \cite{singleton2004learners}.  The father
and the mother used it $67\%$ and $47\%$  of the time, which is $57\%$ of
the total time if Simon's input consists of equal proportions of father's
and mother's speech. Simon's frequency of this incorrect sign is
$73$\%, which is a result of frequency boosting.

This explanation accounts for some
instances where Simon's performance was not closer to the correct ASL
than his parents, without being inconsistent with his frequency
boosting tendency. However, this hypothesis is not enough to explain
all the data.

Correct Secondary Object (SO) morphemes were poorly represented in
Simon's parents' speech. They use the correct forms only $43\%$ and $37\%$
of the time. In contrast to that, Simon's speech contains $59$\% of
the correct variant. That is, Simon is performing frequency boosting
of the variant which is used less that $50\%$ of the time. Another,
and more specific example of this is the particular sign for Plane
(this sign belongs to the CO category), which his parents used
correctly $44\%$ and $11\%$ ($28\%$ on average) of the time, while Simon
boosted this frequency up to $67\%$. Such boosting is not possible if
we use the $M=2$ model of teacher-learner interactions, where there is
only one alternative (incorrect) variant of the rule. If there is
more than one incorrect form, each of which has a lower frequency than
the correct one, then frequency boosting is possible for the correct
variant. 

Another interesting suggestion that apears in the paper is that
Simon's learning happens on a slower scale than that of children
learning from native signers. We have checked this hypothesis and
found that this is indeed consistent with our algorithm. As
demonstrated in figure \ref{fig:con}(b), the speed of convergence of
the algorithm correlates with the source consistency. If the most
frequent variant has a higher value of $\nu$ (the source's dominant
frequency), the convergence of the learner will be higher compared to
relatively lower frequencies of the source's leading variant. This
mechanism suggests that even though Simon was making progress in
frequency boosting his parents' input, it would take him a longer time
to reach the same consistency level than it would take for children learning from
native signers. This is consistent with the suggestion that ``Simon
may still, at age 7, be performing like much younger NN (Native of
Native) children on handshape morphology''.

Finally, the paper suggests that the complexity of the input plays an
important role for the speed of learning and accounts for the lack of
Simon's performance when it comes to certain types of morphemes. In
particular, it is noted that handshape morphemes of the ASL verbs of
motion come in two forms, the simpler semantic classifier and the more
complex size-and-shape specifier (SASS). Simon performs more poorly when it
comes to SASS type morphemes. In our model, the complexity of the
rules may influence the speed at which a rule is learned, if we relate
it to the increment of learning, $s$. We can speculate that more
complex rules are characterized by smaller increments of learning,
thus leading to a larger lag for Simon's speed of learning compared to
that of
the children learning from a consistent input. 

Another way to interpret the influence of rule complexity is to note
that in the example given by the paper, the complex SASS morphemes
consist of several morphemes, one for shape, one for size etc.
Therefore, learning such a morpheme can be represented as learning
several separate rules. If in each of the rules Simon is slightly
below the performance of the NN children of his age, the
multiplicative effect will make his overall scoring for SASS morphemes
even lower.

\section{Conclusions}
\label{sect:disc}

This paper presents a model that explains the frequency boosting
effect observed in children learning a language. It provides a simple
algorithm of the reinforcement type that represents successful
learning without the learner possessing any explicit innate
biases. 

The boosting property of the algorithm was proved analytically in the
case of two forms of the rule, and demonstrated numerically for larger
numbers of forms. Convergence speed and its dependence on the
parameters and the source complexity was also studied.

Finally, the findings were discussed in the context of the study \cite{singleton2004learners}, to demonstrate that the simple model is capable of explaining several key features of Simon's performance and learning behavior.


\begin{thebibliography}{99}

\bibitem[\protect\citename{Andersen, }1983]{andersen1983pidginization}
Andersen, R.~W. (1983).
\newblock {\em Pidginization and Creolization as Language Acquisition.}
\newblock ERIC.

\bibitem[\protect\citename{Brighton, }2002]{brighton2002compositional}
Brighton, H. (2002).
\newblock Compositional syntax from cultural transmission.
\newblock {\em Artificial Life} {\bf 8}(1), 25--54.

\bibitem[\protect\citename{Griffiths \& Kalish, }2007]{griffiths2007language}
Griffiths, T.~L. and Kalish, M.~L. (2007).
\newblock Language evolution by iterated learning with bayesian agents.
\newblock {\em Cognitive Science} {\bf 31}(3), 441--480.

\bibitem[\protect\citename{Hudson~Kam \& Newport,
  }2005]{hudson2005regularizing}
Hudson~Kam, C.~L. and Newport, E.~L. (2005).
\newblock Regularizing unpredictable variation: The roles of adult and child
  learners in language formation and change.
\newblock {\em Language Learning and Development} {\bf 1}(2), 151--195.

\bibitem[\protect\citename{Hudson~Kam \& Newport, }2009]{kam2009getting}
Hudson~Kam, C.~L. and Newport, E.~L. (2009).
\newblock Getting it right by getting it wrong: When learners change languages.
\newblock {\em Cognitive Psychology} {\bf 59}(1), 30.

\bibitem[\protect\citename{Kirby \bgroup \em et al.\egroup ,
  }2004]{kirby2004ug}
Kirby, S., Smith, K., and Brighton, H. (2004).
\newblock From ug to universals: Linguistic adaptation through iterated
  learning.
\newblock {\em Studies in Language} {\bf 28}(3), 587--607.

\bibitem[\protect\citename{Kirby, }1999]{kirby1999function}
Kirby, S. (1999).
\newblock {\em Function, Selection, and Innateness: The Emergence of Language
  Universals: The Emergence of Language Universals}.
\newblock OUP Oxford.

\bibitem[\protect\citename{Kirby, }2001]{kirby2001spontaneous}
Kirby, S. (2001).
\newblock Spontaneous evolution of linguistic structure-an iterated learning
  model of the emergence of regularity and irregularity.
\newblock {\em Evolutionary Computation, IEEE Transactions on} {\bf 5}(2),
  102--110.

\bibitem[\protect\citename{Kroch \& Taylor, }1997]{kroch1997verb}
Kroch, A. and Taylor, A. (1997).
\newblock Verb movement in old and middle english: Dialect variation and
  language contact.
\newblock In: {\em Parameters of morphosyntactic change} pp. 297--325.
  Cambridge: Cambridge University Press.

\bibitem[\protect\citename{Kroch, }1989]{kroch1989reflexes}
Kroch, A. (1989).
\newblock Reflexes of grammar in patterns of language change.
\newblock {\em Language variation and change} {\bf 1}(3), 199--244.

\bibitem[\protect\citename{Lee \bgroup \em et al.\egroup ,
  }2012]{lee2012neural}
Lee, D., Seo, H., and Jung, M.~W. (2012).
\newblock Neural basis of reinforcement learning and decision making.
\newblock {\em Annual review of neuroscience} {\bf 35}, 287--308.

\bibitem[\protect\citename{Maia, }2009]{maia2009reinforcement}
Maia, T.~V. (2009).
\newblock Reinforcement learning, conditioning, and the brain: Successes and
  challenges.
\newblock {\em Cognitive, Affective, \& Behavioral Neuroscience} {\bf 9}(4),
  343--364.

\bibitem[\protect\citename{Marchman \bgroup \em et al.\egroup ,
  }1997]{marchman1997overregularization}
Marchman, V.~A., Plunkett, K., and Goodman, J. (1997).
\newblock Overregularization in english plural and past tense inflectional
  morphology: A response to marcus.
\newblock {\em Journal of Child Language} {\bf 24}, 767--779.

\bibitem[\protect\citename{Marcus \bgroup \em et al.\egroup ,
  }1992]{marcus1992overregularization}
Marcus, G.~F., Pinker, S., Ullman, M., Hollander, M., Rosen, T.~J., Xu, F., and
  Clahsen, H. (1992).
\newblock Overregularization in language acquisition.
\newblock {\em Monographs of the Society for research in child development} pp.
  i--178.

\bibitem[\protect\citename{Marcus, }1995]{marcus1995children}
Marcus, G.~F. (1995).
\newblock Children's overregularization of english plurals: a quantitative
  analysis.
\newblock {\em Journal of Child Language} {\bf 22}, 447--447.

\bibitem[\protect\citename{Norman, }1972]{norman}
Norman, M. (1972).
\newblock {\em Markov Processes and Learning Models}.
\newblock New York: Academic Press.

\bibitem[\protect\citename{Pearl \& Weinberg, }2007]{pearl2007input}
Pearl, L. and Weinberg, A. (2007).
\newblock Input filtering in syntactic acquisition: Answers from language
  change modeling.
\newblock {\em Language Learning and Development} {\bf 3}(1), 43--72.

\bibitem[\protect\citename{Plunkett \& Juola, }1999]{plunkett1999connectionist}
Plunkett, K. and Juola, P. (1999).
\newblock A connectionist model of english past tense and plural morphology.
\newblock {\em Cognitive Science} {\bf 23}(4), 463--490.

\bibitem[\protect\citename{Reali \& Griffiths, }2009]{reali2009evolution}
Reali, F. and Griffiths, T.~L. (2009).
\newblock The evolution of frequency distributions: Relating regularization to
  inductive biases through iterated learning.
\newblock {\em Cognition} {\bf 111}(3), 317--328.

\bibitem[\protect\citename{Sebba, }1997]{sebba1997contact}
Sebba, M. (1997).
\newblock {\em Contact languages: Pidgins and creoles}.
\newblock Macmillan London.

\bibitem[\protect\citename{Senghas \& Coppola, }2001]{senghas2001children}
Senghas, A. and Coppola, M. (2001).
\newblock Children creating language: How nicaraguan sign language acquired a
  spatial grammar.
\newblock {\em Psychological Science} {\bf 12}(4), 323--328.

\bibitem[\protect\citename{Senghas \bgroup \em et al.\egroup ,
  }1997]{senghas1997argument}
Senghas, A., Coppola, M., Newport, E.~L., and Supalla, T.
\newblock Argument structure in nicaraguan sign language: The emergence of
  grammatical devices.
\newblock In: {\em Proceedings of the 21st Annual Boston University Conference
  on Language Development} volume~2 pp. 550--561 1997.

\bibitem[\protect\citename{Senghas, }1995]{senghas1995development}
Senghas, A.
\newblock The development of nicaraguan sign language via the language
  acquisition process.
\newblock In: {\em Proceedings of the 19th Annual Boston University Conference
  on Language Development} pp. 543--552 1995.

\bibitem[\protect\citename{Singleton \& Newport, }2004]{singleton2004learners}
Singleton, J.~L. and Newport, E.~L. (2004).
\newblock When learners surpass their models: The acquisition of american sign
  language from inconsistent input.
\newblock {\em Cognitive Psychology} {\bf 49}(4), 370--407.

\bibitem[\protect\citename{Smith \& Wonnacott, }2010]{smith2010eliminating}
Smith, K. and Wonnacott, E. (2010).
\newblock Eliminating unpredictable variation through iterated learning.
\newblock {\em Cognition} {\bf 116}(3), 444--449.

\bibitem[\protect\citename{Smith \bgroup \em et al.\egroup ,
  }2003]{smith2003iterated}
Smith, K., Kirby, S., and Brighton, H. (2003).
\newblock Iterated learning: a framework for the emergence of language.
\newblock {\em Artificial Life} {\bf 9}(4), 371--386.

\bibitem[\protect\citename{Sutton \& Barto, }1998]{sutton1998reinforcement}
Sutton, R.~S. and Barto, A.~G. (1998).
\newblock {\em Reinforcement learning: An introduction} volume~1.
\newblock Cambridge Univ Press.

\bibitem[\protect\citename{Thomason \& Kaufman, }1991]{thomason1991language}
Thomason, S.~G. and Kaufman, T. (1991).
\newblock {\em Language contact, creolization, and genetic linguistics}.
\newblock Univ of California Press.

\end{thebibliography}

\end{document}